\documentclass[sigconf, nonacm]{acmart}

\setcopyright{none}
\renewcommand\footnotetextcopyrightpermission[1]{}
\usepackage{times}
\usepackage{epsfig}
\usepackage{graphicx}
\usepackage{amsmath}

\usepackage{amssymb}
\usepackage{booktabs}
\usepackage{multirow}
\usepackage{makecell}
\usepackage{adjustbox}
\usepackage{tabularx}
\usepackage{colortbl}
\usepackage{enumitem}

\usepackage{xcolor}
\definecolor{cvprblue}{rgb}{0.21,0.49,0.74}

\usepackage[capitalize,noabbrev]{cleveref}

\usepackage{microtype}
\usepackage{siunitx}
\usepackage{bm}
\usepackage{nicefrac}
\usepackage{comment}

\usepackage{tikz}
\usetikzlibrary{positioning,matrix,arrows.meta,fit,backgrounds,calc}
\usepackage[dvipsnames]{xcolor}
\definecolor{cvprblue}{rgb}{0.21,0.49,0.74}

\usepackage{pifont}

\title{T-IMPACT: A Severity-Aware Benchmark for Contextual Image--Text Manipulation}

\author{Gagandeep Singh}
\affiliation{%
  \institution{The University of Queensland}
  \city{Brisbane}
  \country{Australia}}
\email{gagandeep.singh2@student.uq.edu.au}
\author{Aaditya Yadav}
\affiliation{%
  \city{Brisbane}
  \country{Australia}}

\author{Priyanka Singh}
\affiliation{%
  \institution{The University of Queensland}
  \city{Brisbane}
  \country{Australia}}
\email{priyanka.singh@uq.edu.au}

\begin{abstract}
Recent advances in vision--language models and generative editing systems have made it increasingly easy to produce persuasive multimodal misinformation by altering images, text, or both jointly. However, existing datasets focus mainly on authenticity, out-of-context mismatch, or manipulation type, and rarely capture how strongly an edit changes the likely interpretation of a post. We introduce T-IMPACT, a first-release severity-aware benchmark for manipulated news-style image--text pairs. T-IMPACT contains 98{,}786 examples spanning pristine, image-only, text-only, and joint manipulations, with a calibrated continuous severity signal, coarse low/medium/high labels, and supporting grounding metadata. Starting from a news image--text pair, the pipeline extracts semantic anchors, grounds them spatially, performs localized image edits and constrained caption rewrites, and calibrates contextual-impact scores using limited human ratings. In this release, the calibrated continuous score is the primary severity target, while the low/medium/high bands should be interpreted as coarse operating buckets rather than balanced classes. Experiments show that current models recover some authenticity signal, but severity prediction remains substantially harder and only weakly aligned with human judgment. T-IMPACT provides an initial benchmark for studying multimodal manipulation beyond binary real/fake classification toward graded contextual impact.
\end{abstract}

\ccsdesc[500]{Computing methodologies~Image manipulation}
\ccsdesc[300]{Information systems~Multimedia databases}
\ccsdesc[300]{Computing methodologies~Computer vision}

\keywords{multimodal misinformation, benchmark dataset, cross-modal manipulation, severity assessment, media forensics, vision-language models}

\begin{document}
\maketitle

\section{Introduction}

\newcolumntype{L}[1]{>{\raggedright\arraybackslash}p{#1}}
\newcolumntype{C}[1]{>{\centering\arraybackslash}p{#1}}

\begin{table*}[t!]
\centering
\scriptsize
\setlength{\tabcolsep}{3pt}
\renewcommand{\arraystretch}{0.95}
\begin{tabularx}{\textwidth}{@{}L{2.35cm} C{0.75cm} L{2.0cm} L{2.0cm} L{1.65cm} C{1.1cm} C{1.45cm} C{1.15cm} C{1.45cm} C{1.0cm}@{}}
\toprule
\textbf{Dataset} & \textbf{Year} & \textbf{Domain} & \textbf{Tasks} & \textbf{Manip.\ Types} & \makecell[c]{\textbf{Consis-}\\\textbf{tency}} & \textbf{Grounding} & \makecell[c]{\textbf{Semantic}\\\textbf{Impact}} & \makecell[c]{\textbf{Graded}\\\textbf{Severity}} & \textbf{Scale} \\
\midrule
Fakeddit \cite{nakamura2020fakeddit} & 2020 & Fake news & Multiclass fake news & -- & No & No & No & No & 1M+ \\
Weibo-21 \cite{weibo21} & 2021 & Social media fake news & Fake/Real & -- & No & No & No & No & 9.1K \\
FACTIFY \cite{mishra2021factify} & 2021 & Fact verification & Supp./Ref./NEI & -- & No & No & No & No & 50K \\
FACTIFY3M \cite{chakraborty2023factify3m} & 2023 & Fact verification & QA + fact check & -- & No & No & No & No & 3.0M \\
NewsCLIPpings \cite{luo2021newsclippings} & 2021 & OOC detection & In/out of context & Automatic swaps & Yes & No & Partial & No & $\sim$500K \\
COSMOS \cite{aneja2021cosmos} & 2021 & OOC detection & In/out of context & Curated swaps & Yes & No & Partial & No & 200K / 450K \\
DGM$^4$ \cite{shao2023dgm4} & 2023 & Manipulation detection & Multi-label detection & Several & Yes & Boxes, tokens & No & No & 230K \\
MFND \cite{ijcai2025p891} & 2025 & Fake news + localization & Real/Fake + spatial loc. & 11 types & Yes & Boxes, labels & No & No & 125K \\
CSI-IMD \cite{chen2025csiimd} & 2025 & Image manipulation & Detection + semantics & Several & No & Boxes & Yes & Partial & 502K \\
SIDA / SID-Set \cite{huang2025sida} & 2025 & Deepfake explanation & Detection + localization + explanation & Several & Yes & Saliency / masks & Partial & No & 300K \\
\midrule
\textbf{T-IMPACT (Ours)} & 2026 & Multimodal contextual impact & Authenticity, modality, severity & Remove/Replace/Attr. & Yes & Boxes, masks, tokens & Yes & Yes (L/M/H + score) & 98,786 \\
\bottomrule
\end{tabularx}
\caption{Comparison with representative misinformation, out-of-context, and multimodal manipulation datasets. T-IMPACT differs from prior resources by combining grounded multimodal edits with explicit semantic-impact supervision and an initial graded severity signal.}
\label{tab:related_datasets}
\end{table*}

Synthetic media is no longer confined to research settings. Diffusion models and large language models can now edit images, rewrite captions, and coordinate image--text changes in ways that are realistic, targeted, and easy to scale. As a result, multimodal misinformation is becoming harder to audit: the problem is not only whether content is manipulated, but also how strongly a manipulation changes the meaning of a post \cite{alam2022survey,harris2024fakenews,li2024multimodaldisinfo}.

Most prior work approaches this problem as binary or categorical detection, asking whether a post is real or fake, or whether an image and caption are consistent or inconsistent \cite{nakamura2020fakeddit,mishra2021factify,chakraborty2023factify3m,luo2021newsclippings,aneja2021cosmos,shao2023dgm4,ijcai2025p891}. These formulations are important, but they flatten an important distinction: manipulations do not all matter equally. A small stylistic change may have little effect on interpretation, while a localized visual edit or a short caption rewrite can materially change perceived intent, attribution, legitimacy, or threat. In practice, downstream stakeholders often need to know not only \emph{whether} content was manipulated, but \emph{how much} that manipulation changes the story.

This motivates \emph{severity-aware contextual manipulation}. We use this term to describe manipulations whose impact depends on how image content, textual framing, and surrounding context interact. Severity is therefore not a property of an object or phrase in isolation. The same inserted item, removed region, or rewritten clause can be negligible in one setting and highly consequential in another, depending on scene context, actor roles, and the narrative implied by the image--text pair. This is especially important in multimodal posts, where meaning is jointly constructed across modalities rather than carried by image or text alone.

Existing benchmarks rarely model this graded impact directly. Most provide binary authenticity labels, coarse manipulation categories, or localization annotations \cite{alam2022survey,harris2024fakenews,li2024multimodaldisinfo}. Recent resources such as MFND, DGM$^4$, and CSI-IMD expand coverage of multimodal manipulation and localization \cite{ijcai2025p891,shao2023dgm4,chen2025csiimd}, but they still do not provide a unified framework for localized multimodal edits paired with a calibrated severity signal that reflects perceived narrative impact across image and text.

To address this gap, we present \textbf{T-IMPACT}, a first-release severity-aware multimodal dataset framework and benchmark for news-style image--text pairs. T-IMPACT is organized around two axes: \emph{modality of manipulation} and \emph{severity of impact}. Along the modality axis, the dataset includes pristine pairs, image-only manipulations, text-only manipulations, and joint image--text manipulations. Along the severity axis, each manipulated pair is assigned a calibrated continuous severity score and a corresponding coarse low/medium/high label derived from a structured scoring pipeline and limited human calibration. The scoring formulation combines object-tier priors, contextual incongruity, multimodal contradiction, visual salience, and visibility of the manipulated region, allowing severity to approximate contextual impact rather than provenance alone. In this release, we treat the calibrated continuous score as the primary severity target, while the low/medium/high labels serve as coarse operating buckets rather than balanced semantic classes.

A key design choice in T-IMPACT is to generate manipulations from localized semantic anchors instead of treating the image or caption as an undifferentiated whole. The pipeline first extracts literal image descriptions and candidate anchors, grounds them spatially, segments editable regions, proposes localized edit plans and headline rewrites, and then executes image edits using inpainting-based generation. This design preserves fine-grained provenance about what was changed, where it was changed, and how the resulting example should be interpreted. The final dataset package also retains auxiliary metadata such as edit type, severity components, quality-control fields, and provenance traces, supporting both benchmarking and dataset auditing. A key limitation of binary authenticity prediction is that it assumes manipulated media can always be meaningfully separated from authentic media as a yes/no decision. However, as generation and editing systems improve, manipulations may become increasingly difficult to distinguish reliably at the artifact level, even when their effect on audience interpretation remains substantial. In such settings, severity modeling becomes especially important: the central question is not only whether a post is manipulated, but how much that manipulation changes the story it appears to tell and how likely it is to influence downstream judgment. T-IMPACT is intended to support this framing by benchmarking contextual impact alongside, rather than in place of, binary authenticity.

\subsection{Contributions}

We make three contributions:

\begin{itemize}
    \item \textbf{Severity-aware contextual manipulation as a dataset objective.}
    We formalize multimodal manipulation severity as the degree to which an edit changes the likely interpretation of a news-style image--text pair, moving beyond binary authenticity toward graded contextual impact.

    \item \textbf{A localized multimodal generation pipeline for T-IMPACT.}
    We construct manipulated examples through anchor extraction, open-vocabulary grounding, segmentation-guided region selection, localized image editing, and paired text rewriting, producing pristine, image-only, text-only, and joint image--text variants with edit-level metadata.

    \item \textbf{An initial benchmark release with calibrated severity metadata.}
    We release a large-scale T-IMPACT build with a calibrated continuous severity signal, coarse low/medium/high severity labels, modality and edit-type annotations, and supporting provenance and quality metadata, enabling evaluation of both manipulation detection and graded impact prediction.
\end{itemize}
\section{Related Work}
\begin{figure*}[t!]
    \centering
    \includegraphics[width=\textwidth]{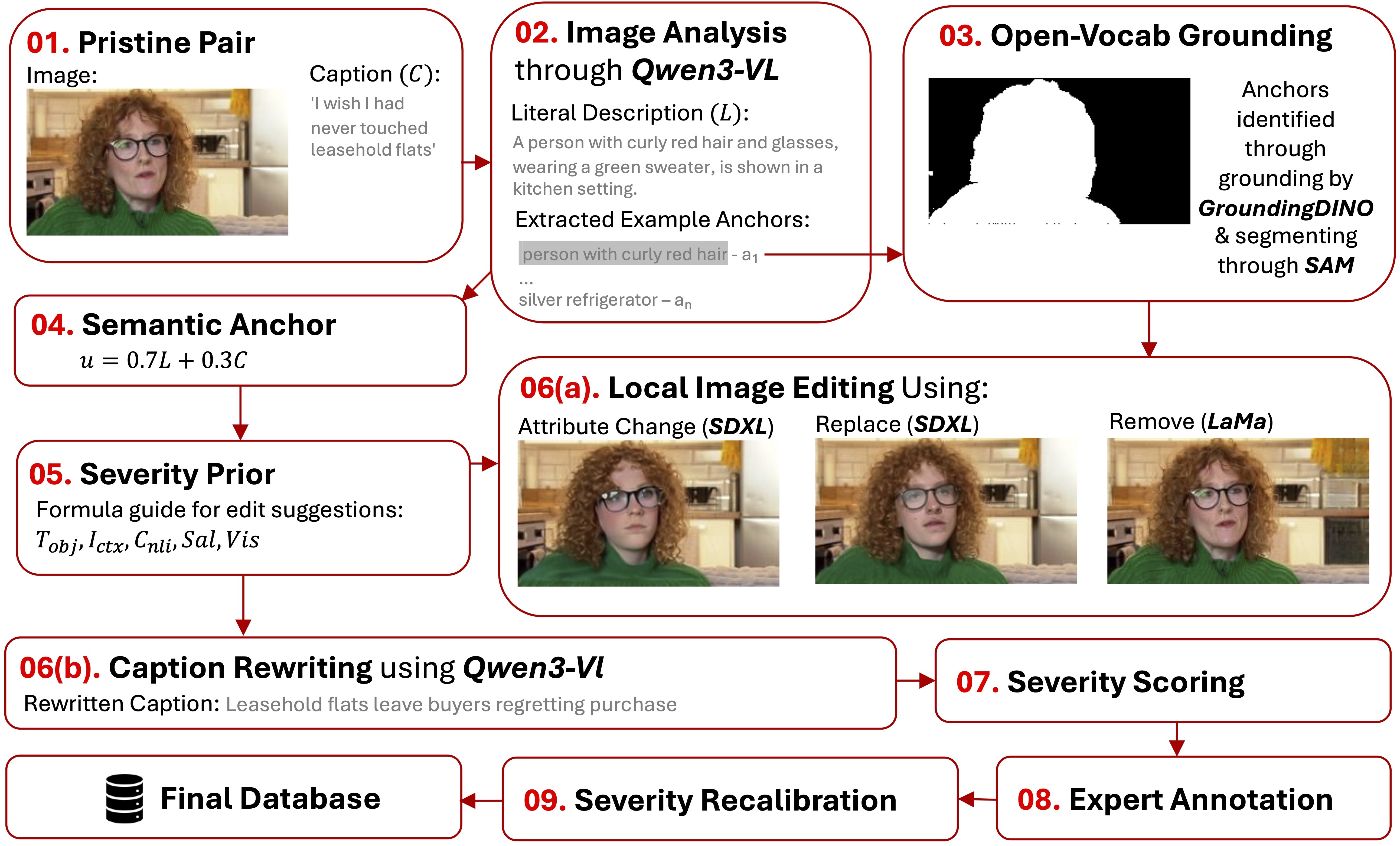}
    \caption{Overview of the T-IMPACT construction pipeline. Starting from a pristine image--text pair, the pipeline extracts semantic anchors, grounds them spatially, performs localized image edits and caption rewrites, computes severity scores, and calibrates them using expert annotations before final packaging.}
    \label{fig:pipeline}
\end{figure*}

Research on misinformation and manipulation spans textual fake news detection, multimodal fact verification, out-of-context media, and localized forensics. We focus on work most relevant to severity-aware multimodal manipulation.

Early fake news research largely emphasized textual signals such as stance, writing style, and propagation behaviour \cite{zafarani2019fakenews,alam2022survey}. While these resources support fine-grained supervision, they do not model accompanying imagery or the magnitude of narrative change. Multimodal benchmarks such as Fakeddit, FACTIFY, and FACTIFY3M extend verification to image--text pairs, while recent surveys show that multimodal systems still center on truthfulness, support/refute decisions, or coarse manipulation labels rather than graded impact \cite{nakamura2020fakeddit,mishra2021factify,chakraborty2023factify3m,li2024multimodaldisinfo,harris2024fakenews}. MFND moves further by adding manipulation-type and localization annotations, but its objective remains detection-oriented rather than severity-oriented \cite{ijcai2025p891}.

A closely related line of work studies misleading reuse of authentic content. NewsCLIPpings and COSMOS benchmark out-of-context detection by pairing real images with misleading captions or narratives, requiring reasoning over image--text compatibility rather than artifact detection alone \cite{luo2021newsclippings,aneja2021cosmos}. MM-Inconsistency similarly frame the problem as contradiction or entailment across modalities. These datasets are important because they move toward contextual harm, but they still use binary or categorical supervision and do not distinguish mild reframing from severe narrative inversion \cite{mmic}.

Recent multimodal forensics datasets add evidence grounding. DGM$^4$ and MFND provide manipulated image--text pairs with spatial and token-level supervision, enabling models to detect manipulations and localize the evidence \cite{shao2023dgm4,ijcai2025p891}. Parallel image-forensics work examines whether a manipulated region is semantically important rather than only visually altered. CSI-IMD is especially relevant because it introduces semantic significance, while SIDA and M-SegEval further emphasize localization and meaningful region assessment \cite{chen2025csiimd,huang2025sida,msegeval}. However, these works are either image-only, explanation-oriented, or region-ranking based; they do not provide a unified multimodal setting with explicit low/medium/high severity labels.

The main gap is therefore not the absence of multimodal data, manipulation types, or grounding in isolation. It is the absence of a benchmark that combines localized image--text manipulation with explicit, context-sensitive severity supervision. Surveys on multimodal disinformation repeatedly note the mismatch between current benchmark labels and the needs of human analysts, journalists, and policymakers, who must triage content by likely impact rather than confidence alone \cite{alam2022survey,li2024multimodaldisinfo,harris2024fakenews}. This concern also aligns with broader work on deepfakes and the liar's dividend, which shows that uncertainty about authenticity can itself become politically useful \cite{chesney2019deepfakes}. T-IMPACT is designed around this missing layer: modality-aware, localized, and severity-graded multimodal manipulation.

\begin{figure*}[t]
    \centering
    \includegraphics[width=\textwidth]{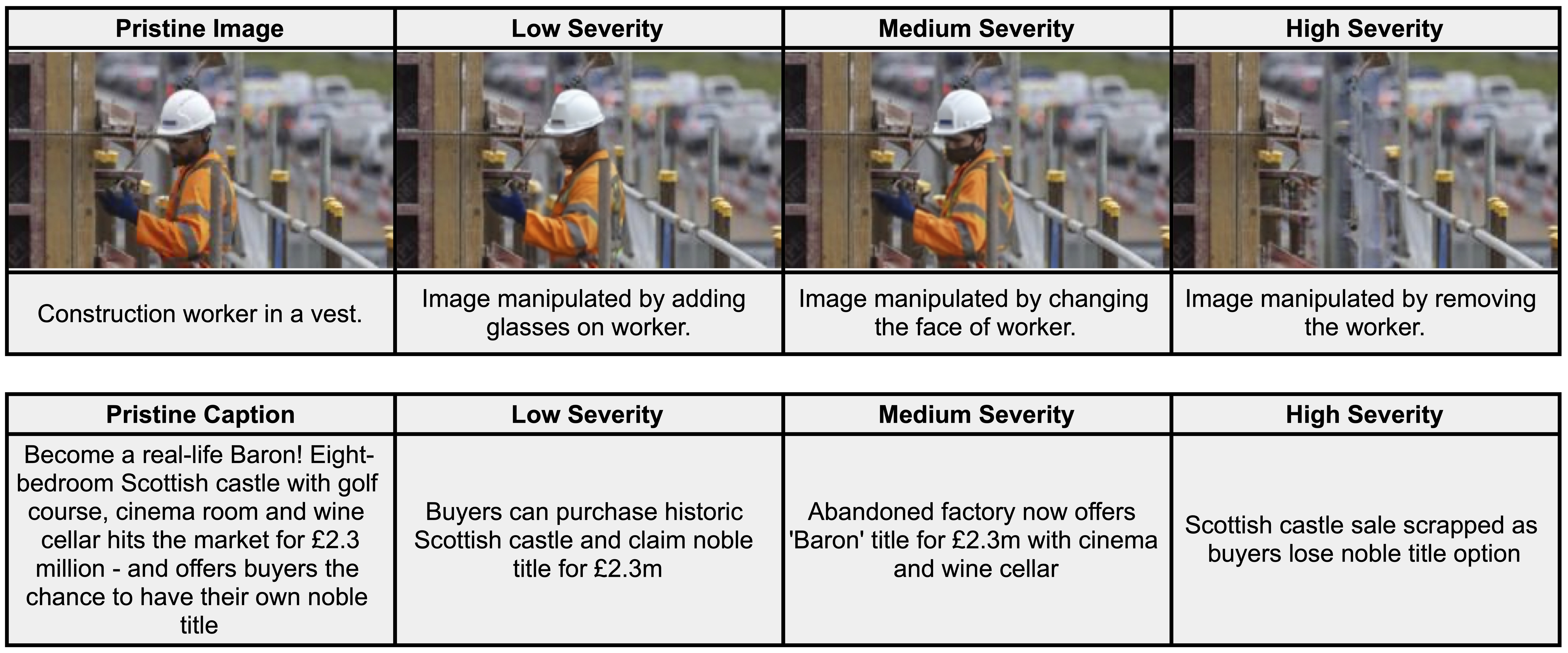}
    \caption{Examples of severity progression in T-IMPACT. The top row shows localized visual manipulations of increasing impact, while the bottom row shows caption rewrites that progressively change the likely interpretation of the original post.}
    \label{fig:severity_examples}
\end{figure*}
\section{Dataset Construction and Overview}
\label{sec:dataset}

T-IMPACT is a first-release severity-aware benchmark for manipulated news-style image--text pairs. Each example contains an image, a headline or short caption, and metadata describing authenticity, manipulation modality, severity, edit type, and supporting evidence. Pristine pairs are derived from the public \emph{News Dataset with Images},\footnote{Available at \url{https://www.kaggle.com/datasets/mdkabinhasan/news-dataset-with-images/data}, accessed April 2026. The collection is described as an MIT-licensed aggregation of RSS-scraped international news articles with associated images. We use it as a retrieval layer for public news pages and preserve source-side metadata for traceability.} and we preserve source-side metadata such as article URL, publication time, and image URL for traceability. The current packaged release contains 98{,}786 examples: 9{,}656 pristine pairs, 48{,}677 image-only manipulations, 35{,}357 text-only manipulations, and 5{,}096 joint image--text manipulations. Joint image--text manipulations therefore form a smaller but explicit portion of the release, reflecting the added constraint that both modalities remain coherent after coordinated editing.

The pipeline begins with anchor-centric semantic extraction. For each pristine image--caption pair, a Qwen vision--language model \cite{qwen25vl} produces a literal visual description, visible text, and a set of localized anchors, attributes, and relations. The prompts explicitly discourage speculation and favor literal, editable entities such as people, objects, signs, devices, and scene regions. A recovery pass then adds missed anchors when the first extraction is incomplete. These outputs are grounded spatially using GroundingDINO \cite{liu2024groundingdino}, which converts anchor phrases into candidate bounding boxes, and refined into object masks using SAM \cite{kirillov2023segmentanything}. The resulting Qwen, grounding, and segmentation outputs are merged into per-sample manifests that preserve the image caption, article-context rewrite, grounded anchor structure, provenance, and mask statistics. This design is central to T-IMPACT: manipulations are tied to explicit semantic anchors rather than applied indiscriminately to the whole image or caption.

Manipulated variants are generated along two axes: modality and severity. On the image side, the planner proposes localized edits for each anchor and severity tier using structured instructions of the form \texttt{[MASK]}, \texttt{[FILL]}, and \texttt{[VISUAL]}. The operations used in the final pipeline are remove, replace, fill/insert, and attribute change. Image execution is localized: LaMa is used for removal and background completion \cite{suvorov2022lama}, while SDXL inpainting is used for insertions, replacements, and attribute edits \cite{podell2023sdxl}. The editing code further adapts edit strength by operation and severity, tightens masks to reduce background bleed, and checks whether the masked region changed enough to be visually meaningful. On the text side, Qwen also produces constrained headline rewrites tied to the same anchor, operation, modality mode, and requested severity, allowing the dataset to support text-only and coordinated image--text manipulations without free-form article rewriting.

Severity is modeled as contextual impact rather than simple artifact presence. For each manipulated pair, the raw score is recomputed from five components:
{\small
\begin{equation}
\begin{aligned}
S_{\mathrm{raw}}= {} .35\,T_{\mathrm{obj}} + .25\,I_{\mathrm{ctx}} + .20\,C_{\mathrm{nli}} + .15\,\mathrm{Sal} + .05\,(1-\mathrm{Vis})
\end{aligned}
\label{eq:sraw}
\end{equation}
}
where $T_{\mathrm{obj}}$ is an object-tier prior, $I_{\mathrm{ctx}}$ measures contextual incongruity, $C_{\mathrm{nli}}$ captures contradiction, $\mathrm{Sal}$ measures salience, and $\mathrm{Vis}$ captures visibility. The weights are used as a structured prior over contextual impact rather than as a fully learned optimum: object importance and contextual incongruity receive the largest influence because they most directly capture narrative change, contradiction and salience provide supporting semantic and perceptual evidence, and visibility is intentionally down-weighted because conspicuity alone does not determine interpretive impact. Further information about this formula can be found in the supplementary materials. A monotonic calibration map derived from limited human ratings is then applied to better align automatic scores with perceived severity \cite{zadrozny2002transforming}. In the released build, the final severity bins are assigned after calibration using thresholds of low $\leq 0.5$, medium $\leq 0.833333$, and high otherwise.

The released benchmark stores severity both as a continuous calibrated score and as a coarse low/medium/high label.The  pre-calibration generation mix, has samples at 20{,}741 low, 37{,}074 medium, and 31{,}315 high.

Final packaging is performed by the human-calibrated release builder, which combines manifests, edit plans, edit results, and text rewrites into \texttt{all\_examples.jsonl}, CSV metadata, split files, a quarantine file, and summary statistics. The public release is partitioned deterministically into train/val/test splits of 79{,}101 / 10{,}039 / 9{,}646 examples. The repository endpoint for code, schema documentation, split definitions, and supplementary materials is \url{https://github.com/Gaganx0/T-IMPACT_dataset}. Code and release tooling are distributed under the MIT License, while T-IMPACT annotations, metadata, manifests, and benchmark packaging created by the authors are released under CC BY-NC 4.0. Underlying third-party news articles and images are not relicensed and remain subject to their original source terms.

\textbf{Ethical considerations.}
T-IMPACT is intended for research on multimodal misinformation detection, calibration, and robustness analysis, not for generating deceptive content. The benchmark is derived from publicly accessible news pages and preserves source metadata for traceability, but we do not claim ownership of the underlying third-party articles or images, which remain subject to their original source terms. The released benchmark excludes flagged sensitive cases and does not provide a turnkey interface for generating deceptive media at scale. We distinguish between benchmark artifacts created in this work and underlying third-party news content: annotations, metadata, manifests, and release packaging are shared under the project’s stated terms, while original source articles and images remain governed by their respective source rights and terms of use.
\section{Evaluation and Discussion}
\label{sec:eval}

\begin{table*}[t]
\centering
\scriptsize
\setlength{\tabcolsep}{3.5pt}
\renewcommand{\arraystretch}{0.92}
\caption{First-release benchmark snapshot on T-IMPACT. Text, fusion-pruned, and Florence-2 are severity-only cross-validation baselines, so authenticity, type, and localization fields are unavailable. Qwen uses hard authenticity outputs for AUROC. TRUST-VL values are $n$-weighted aggregates of the uploaded phase summaries. Upward and downward rates denote predictions above or below the gold severity label. HAMMER is reported only on the tasks exposed by the current evaluation interface, so its row is partial.}
\label{tab:main_shift_merged}
\resizebox{0.98\textwidth}{!}{%
\begin{tabular}{lcccccccccc}
\toprule
Model & F1 $\uparrow$ & AUROC $\uparrow$ & Type F1 $\uparrow$ & MAE $\downarrow$ & $\rho$ $\uparrow$ & QWK $\uparrow$ & Up $\uparrow$ & Down $\uparrow$ & Token F1 $\uparrow$ \\
\midrule
Text baseline (CV) & \textemdash & \textemdash & \textemdash & 0.436 & 0.347 & 0.320 & 0.189 & 0.225 & \textemdash \\
Fusion-pruned baseline (CV) & \textemdash & \textemdash & \textemdash & 0.409 & 0.312 & 0.291 & 0.179 & 0.211 & \textemdash \\
Florence-2 (CV) & \textemdash & \textemdash & \textemdash & 0.483 & 0.252 & 0.216 & 0.261 & 0.167 & \textemdash \\
Qwen & 0.604 & 0.521 & 0.049 & 1.532 & 0.006 & 0.001 & 0.437 & 0.548 & \textemdash \\
TRUST-VL & 0.546 & 0.524 & 0.081 & 0.347 & 0.075 & 0.023 & 0.050 & 0.948 & \textemdash \\
HAMMER & 0.641 & 0.459 & 0.197 & \textemdash & \textemdash & \textemdash & \textemdash & \textemdash & 0.078 \\
LLaVA & 0.562 & 0.526 & 0.261 & 0.326 & 0.121 & 0.001 & 0.009 & 0.991 & \textemdash \\
\bottomrule
\end{tabular}%
}
\end{table*}

\subsection{Experimental setup}

We evaluate T-IMPACT along four axes: \emph{authenticity detection}, \emph{manipulation-type prediction}, \emph{severity prediction}, and \emph{localization} when grounded outputs are available. Table~\ref{tab:human_calibration_small} reports severity calibration against human judgment, and Table~\ref{tab:main_shift_merged} provides a first-release benchmark snapshot across available models and auxiliary baselines. The comparison is intentionally heterogeneous: the text, Florence-2, and fusion-pruned rows are 5-fold CORAL ordinal baselines over precomputed features \cite{cao2019coral}; Qwen3VL-8B is evaluated with short structured prompting \cite{qwen25vl}; TrustVL-13B is evaluated from post-parsed free-form generations \cite{trustvl}; LLaVA-OV-Qwen2-7B-OV is evaluated with task-specific structured prompting \cite{llavaonevision}; and HAMMER is included only on the tasks exposed by the current interface. We therefore use the table as a benchmark snapshot rather than a strict like-for-like leaderboard. The goal is not to claim a final ranking across mismatched model families, but to test whether current systems recover T-IMPACT’s core target: ordered contextual impact.

\subsection{Human calibration and severity difficulty}

Human annotation was collected on a common-rated pool of 500 manipulated samples from 3 raters. Average ordinal agreement is 0.334, exact agreement is 0.172, and adjacent agreement is 0.465, indicating an ordinal but boundary-sensitive task. Raters often agree on a broad impact region while differing between neighboring levels, highlighting the inherent ambiguity of severity judgments in multimodal misinformation.

Table~\ref{tab:human_calibration_small} shows that isotonic calibration improves the automatic severity signal, but only modestly: relative to the raw score, Spearman rises from 0.025 to 0.115, QWK from $-0.012$ to 0.035, and MAE falls from 0.330 to 0.258. However, the calibrated score remains far below the leave-one-rater-out human ceiling ($\rho=0.428$, QWK$=0.383$). This is important for interpretation. The lower MAE should not be read as evidence that severity has been solved, because in this release MAE is helped by compression toward the dominant middle band, whereas Spearman and QWK better expose the remaining weakness in ordinal structure. The calibration study therefore supports the paper’s main framing: severity is not just another label on top of authenticity, but a harder and less stable target that benefits from explicit benchmarking.

\subsection{Benchmark snapshot: authenticity is easier than severity}

The main pattern in Table~\ref{tab:main_shift_merged} is consistent across model families: several models recover some manipulated-versus-pristine signal, but much weaker alignment is observed for ordered impact. Qwen achieves stronger authenticity performance (F1 0.604, AUROC 0.521), yet its severity behavior is extremely weak (MAE 1.532, $\rho=0.006$, QWK 0.001). In other words, Qwen can separate some manipulated samples from pristine ones without preserving the ordinal structure of low, medium, and high contextual impact. Under short structured prompting, it behaves more like a compact classifier than a calibrated severity reasoner.

TrustVL is more stable. Although its authenticity F1 (0.546) is lower than Qwen’s, it is slightly better on AUROC (0.524), better on manipulation type (0.081), and much better on severity MAE (0.347). Even so, its rank-based severity metrics remain weak ($\rho=0.075$, QWK$=0.023$). The right reading is therefore not that TrustVL solves severity, but that it is less erratic than Qwen while still failing to recover the benchmark’s ordinal geometry reliably.

LLaVA shows a related but slightly stronger severity pattern. LLaVA-OV-Qwen2-7B-OV reaches competitive authenticity performance (F1 0.562, AUROC 0.526), gives the strongest manipulation-type result among the fully populated VLM rows (0.261), and achieves the best severity MAE among the prompted multimodal models (0.326). However, its ordinal fidelity remains poor ($\rho=0.121$, QWK 0.001). This distinction matters. LLaVA is numerically closer to the gold label on average, but it still does not preserve the ordered relationships between severity levels in a human-aligned way. T-IMPACT is useful precisely because it separates these two notions rather than collapsing them into a single score.

HAMMER should be interpreted separately because its row is partial. Within the currently evaluated tasks, it gives the strongest authenticity F1 (0.641), stronger type prediction than Qwen and TRUST-VL (0.197), and the only reported token-level localization signal (Token F1 0.078). At the same time, its AUROC is weak (0.459), and no directly comparable severity outputs are available here. The present evidence therefore supports HAMMER as a useful manipulation-sensitive detector with localized evidence, but not yet as a directly comparable severity model. This distinction is important for the narrative of the paper: localization and manipulation typing are valuable, but they do not by themselves answer how strongly a manipulation changes likely interpretation.

\subsection{Severity-only baselines and what they reveal}

The severity-only baselines are informative because they isolate the ordinal prediction problem from the open-ended prompting behavior of large VLMs. The text baseline gives the strongest ordinal consistency ($\rho=0.347$, QWK$=0.320$), while the fusion-pruned baseline gives the best MAE (0.409). Florence-2 is weaker on both axes \cite{florence2}. Two conclusions follow.

First, in the current release, severity appears to be encoded more reliably in controlled textual reframing than in generic visual features alone. This is plausible given how the dataset is constructed: many manipulations are designed to shift narrative interpretation through localized semantic change, and the corresponding caption rewrite often expresses that shift more directly than broad visual embeddings do.

Second, adding visual features does not automatically improve severity prediction. Relative to text-only, the fusion-pruned baseline reduces average distance to the gold label but loses rank structure. That is exactly the kind of trade-off a severity-aware benchmark should expose. A model can look better under one metric while still failing to preserve the ordinal relationships that matter for impact-aware judgment.

\subsection{Directional errors and benchmark value}

Directional errors make these differences clearer. Qwen has both high upward and downward shift rates (0.437 and 0.548), indicating unstable severity behavior rather than a simple conservative or aggressive bias. TrustVL is strongly downward biased (0.050 up, 0.948 down), and LLaVA is even more so (0.009 up, 0.991 down). This helps explain why lower MAE does not translate into stronger ordinal agreement: both models compress predictions toward safer, lower-severity judgments.

This directional view is one of the practical advantages of T-IMPACT. A system that systematically under-calls severe manipulations is operationally different from one that over-flags milder edits, even if their average absolute errors are similar. Standard fake/real evaluation would largely miss this distinction. By separating whether a manipulation is detected from whether its likely impact is estimated in the correct direction and at roughly the correct level, T-IMPACT exposes a failure mode that authenticity benchmarks typically hide.

Overall, the benchmark snapshot suggests a clear conclusion: severity-aware multimodal judgment is distinct from, and substantially harder than, authenticity detection. Authenticity, type prediction, and localization each capture part of the problem, but none of them alone implies reliable impact-aware interpretation.

\begin{table}[t]
\centering
\scriptsize
\setlength{\tabcolsep}{4pt}
\renewcommand{\arraystretch}{0.95}
\caption{Severity score calibration against human judgment. The calibration analysis uses 500 manipulated items to form the annotation pool. Human ceiling is leave-one-rater-out against the mean of the other two raters.}
\label{tab:human_calibration_small}
\resizebox{0.98\columnwidth}{!}{%
\begin{tabular}{lccc}
\toprule
Method / Subset & Spearman $\rho$ $\uparrow$ & QWK $\uparrow$ & MAE $\downarrow$ \\
\midrule
Raw severity score & 0.025 & -0.012 & 0.330 \\
Calibrated severity score & 0.115 & 0.035 & 0.258 \\
Human ceiling & 0.428 & 0.383 & 0.291 \\
\bottomrule
\end{tabular}%
}
\end{table}
\section{Conclusion}
\label{sec:conclusion}

T-IMPACT introduces a first-release benchmark for severity-aware contextual image--text manipulation. Unlike prior datasets that focus primarily on authenticity, out-of-context mismatch, or localization, T-IMPACT is designed to evaluate how strongly a manipulation changes the likely interpretation of a post. The current results suggest that this is a distinct and substantially harder problem than authenticity detection: several models recover some manipulated-versus-pristine signal, but much weaker alignment is observed for ordered impact.

Accordingly, the present release is best interpreted through two complementary severity views. The primary target is a calibrated continuous impact score, which is used throughout the benchmark to capture graded contextual change. The accompanying low/medium/high labels are retained as coarse operating buckets rather than as a balanced three-class prediction target. 

These observations define the most important next steps. First, future releases should strengthen severity supervision through more annotations, more overlapping raters, and larger common-rated calibration subsets. Second, the thresholding and resampling strategy should be revisited so that the coarse severity buckets better reflect the dynamic range of the underlying continuous signal. Third, T-IMPACT can be expanded with additional edit families, including more diverse synthetic and multimodal inconsistencies. Finally, future models should combine localization with explicit severity prediction so that they estimate not only whether and where a manipulation occurs, but also how much it changes likely interpretation. Overall, T-IMPACT is intended as an initial benchmark for \emph{contextual impact}, not just authenticity.

\bibliographystyle{ACM-Reference-Format}
\bibliography{references}

\end{document}